\documentclass[letterpaper, 10 pt, conference]{ieeeconf}   

\IEEEoverridecommandlockouts                              
\usepackage[utf8]{inputenc}
\usepackage[T1]{fontenc}

\usepackage{import}
\usepackage{amsmath}
\usepackage{amsfonts}
\usepackage{graphicx}
\usepackage{caption}
\usepackage{subcaption}
\usepackage{booktabs}
\usepackage{multirow}
\usepackage{tikz}
\usepackage{tikz/tkz-kiviat}
\usetikzlibrary{arrows}
\usepackage{tcolorbox}
\usepackage{diagbox}
\usepackage{cite}
\usepackage{bm}
\usepackage{siunitx}
\usepackage{svg}
\usepackage{xcolor}
\usepackage{arydshln}
\usepackage{url}
\usepackage{hyperref}

\definecolor{mplblue}{HTML}{1f77b4}
\definecolor{mplorange}{HTML}{ff7f0e}
\definecolor{mplgreen}{HTML}{2ca02c}
\definecolor{mplred}{HTML}{d62728}
\definecolor{mplpurple}{HTML}{9467bd}
\definecolor{mplbrown}{HTML}{8c564b}

\makeatletter

\title{\LARGE \bf
Neural Rendering for Sensor Adaptation in 3D Object Detection
}

\author{Felix Embacher$^{*1,2}$, David Holtz$^{*1,3}$, Jonas Uhrig$^{1}$, Marius Cordts$^{1}$, Markus Enzweiler$^{2}$ 
\thanks{*These authors contributed equally to this work.}
\thanks{$^{1}$Mercedes-Benz AG, \{felix.embacher, david.holtz\}@mercedes-benz.com
}%
\thanks{$^{2}$Institute for Intelligent Systems, Esslingen University of Applied Sciences}%
\thanks{$^3$University of Stuttgart}
}
\newcommand\copyrighttext{%
  \footnotesize \textcopyright 2025 IEEE. Personal use of this material is permitted.
  Permission from IEEE must be obtained for all other uses, in any current or future
  media, including reprinting/republishing this material for advertising or promotional
  purposes, creating new collective works, for resale or redistribution to servers or
  lists, or reuse of any copyrighted component of this work in other works.}
\newcommand\copyrightnotice{%
\begin{tikzpicture}[remember picture,overlay]
\node[anchor=south,yshift=10pt] at (current page.south) 
  {\fbox{\parbox{\dimexpr\textwidth-\fboxsep-\fboxrule\relax}{\copyrighttext}}};
\end{tikzpicture}%
}
\begin{document}

\maketitle
\copyrightnotice
\thispagestyle{empty}
\pagestyle{empty}

\begin{abstract}
    Autonomous vehicles often have varying camera sensor setups, which is inevitable due to restricted placement options for different vehicle types. Training a perception model on one particular setup and evaluating it on a new, different sensor setup reveals the so-called cross-sensor domain gap, typically leading to a degradation in accuracy. In this paper, we investigate the impact of the cross-sensor domain gap on state-of-the-art 3D object detectors. To this end, we introduce CamShift, a dataset inspired by nuScenes and created in CARLA to specifically simulate the domain gap between subcompact vehicles and sport utility vehicles (SUVs). Using CamShift, we demonstrate significant cross-sensor performance degradation, identify robustness dependencies on model architecture, and propose a data-driven solution to mitigate the effect.
    
    On the one hand, we show that model architectures based on a dense Bird's Eye View (BEV) representation with backward projection, such as BEVFormer, are the most robust against varying sensor configurations. On the other hand, we propose a novel data-driven sensor adaptation pipeline based on neural rendering, which can transform entire datasets to match different camera sensor setups. Applying this approach improves performance across all investigated 3D object detectors, mitigating the cross-sensor domain gap by a large margin and reducing the need for new data collection by enabling efficient data reusability across vehicles with different sensor setups.

    The CamShift dataset and the sensor adaptation benchmark are available at \href{https://dmholtz.github.io/camshift/}{https://dmholtz.github.io/camshift/}.
\end{abstract}
\section{INTRODUCTION}
    Visual perception is crucial for autonomous vehicles to understand their surroundings and to make reasonable driving decisions. Although camera-based 3D object detectors are already based on a significantly less complex sensor setup compared to LiDAR-based and multimodal approaches, scaling the methods to different autonomous vehicles remains challenging. Even with identical hardware and an equal number of cameras, the mounting positions of the cameras still vary across vehicles. As a result, the same perception models cannot necessarily be rolled out to an entire vehicle fleet, as the cross-sensor domain gap between different vehicle types can negatively affect perception accuracy.
    
\begin{figure}[t]
    \centering
    \begingroup
    \footnotesize
    \includesvg[width=0.48\textwidth]{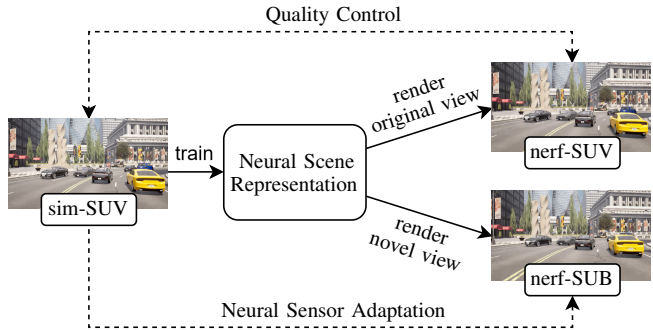}
    \endgroup
    \caption{We propose a novel sensor adaptation pipeline based on neural rendering to convert entire datasets from one sensor setup to another. We train a neural scene representation for each sequence of surrounding camera images captured with SUV cameras, render the original view for quality control, and finally render the novel view of a subcompact vehicle.}
    \label{fig:pipeline}
\end{figure}

    This raises two important open research questions. Firstly, how robust are camera-based perception models to the cross-sensor domain gap? And secondly, can we mitigate the cross-sensor degradation with a data-driven approach? We address these questions by designing a novel sensor adaptation benchmark to specifically investigate the cross-sensor domain gap between different sensor setups, demonstrated via camera-based 3D object detection as perception task. Currently, there is no available dataset that isolates the cross-sensor domain gap, enabling dedicated analysis of sensor adaptation from one vehicle type to another. In this work, we propose the CamShift dataset, consisting of two splits (\texttt{sim-SUB} for subcompact vehicles and \texttt{sim-SUV} for SUVs) that are simulated simultaneously using the open-source urban driving simulator CARLA \cite{carladosovitskiy2017}. Using a simulator allows for a controlled evaluation framework, while the assessed approaches and the insights obtained can be directly applied to real-world situations. We assess the impact of cross-sensor degradation on representative camera-based 3D object detectors by training models on one sensor setup and evaluating their detection accuracy on the other setup with a different perspective. We find that a model's ability to adapt to different sensor setups depends heavily on its representation and processing approach. Specifically, methods that are sparse and object-centric behave differently from those that use dense BEV-based representations. As a further step to mitigate this cross-sensor degradation, we propose a novel data-driven sensor adaptation pipeline based on Neural Radiance Fields (NeRFs) \cite{nerfmildenhall2021} to transform whole datasets to new sensor setups (see Figure \ref{fig:pipeline}). We demonstrate the effectiveness of our data-driven sensor adaptation pipeline to significantly close the cross-sensor domain gap, thereby enabling data reusability in the development of autonomous vehicles.

    The main contributions of this paper are:
    \begin{itemize}
        \item We develop a novel sensor adaptation benchmark to exclusively examine the cross-sensor domain gap between different vehicle types in 3D object detection.
        \item We conduct a systematic study of representative camera-based 3D object detectors when subjected to the cross-sensor domain gap, and find that dense BEV-based methods with backward projection are the most robust.
        \item We derive a novel data-driven sensor adaptation pipeline based on neural rendering and demonstrate its effectiveness in significantly closing the cross-sensor domain gap.
    \end{itemize}
\section{RELATED WORK}
\subsection{Camera-based 3D Object Detection}
In the field of autonomous driving, there is an increasing trend towards camera-only 3D object detection to enable cost-effective perception. Yet, to achieve a comprehensive understanding of the entire environment, camera-only methods rely on multi-view images and temporal aggregation to fuse information from previous timestamps. As shown in Table \ref{tab:3d_object_detection_architectures}, both single-frame and temporal approaches can be categorized into two types depending on their intermediate representation.
\subsubsection{Dense BEV-based 3D Object Detection}
Typically, dense BEV-based 3D object detectors are built on a BEV encoder that processes multi-view camera images to extract dense, grid-shaped BEV features. Subsequently, the 3D object detection task is performed based on these BEV features. Well-known dense BEV-based methods are the BEVDet variants \cite{bevdethuang2021, bevdet4dhuang2022, bevpoolv2huang2022} and BEVFormer \cite{bevformerLi2022}. While BEVDet uses the forward projection of LSS \cite{lssphilion2020} to lift 2D features to the BEV space, BEVFormer uses backward projection based on the camera parameters to project 3D reference points to the 2D image space. Furthermore, BEVFormer introduces the spatial cross-attention mechanism based on deformable attention \cite{deformableDetrzhu2020} to extract information from the multi-view 2D image features. While BEVDet4D \cite{bevdet4dhuang2022} extends BEVDet into a temporal 3D object detector, BEVFormer-S \cite{bevformerLi2022} serves as the static, non-temporal variant of BEVFormer.
\subsubsection{Sparse Object-centric 3D Object Detection}
In contrast to dense BEV-based methods, queries in sparse object-centric methods correspond to a set of object predictions and can directly interact with 2D image features. This enables 3D perception tasks without any dense intermediate representations \cite{surveyVisionCentricBEVPerceptionma2024}. DETR3D \cite{detr3dwang2022} is a fundamental 3D object detection framework that relies on the geometric back-projection of 3D reference points using camera matrices. In each iteration, DETR3D decodes the 3D bounding box center from each query and projects it into all image planes where it could be visible. It then samples features at the projected points and uses them to update each object query. Unlike DETR3D, PETR \cite{petrliu2022} and SpatialDETR \cite{spatialdetrdoll2022} refine object queries by introducing a 3D position encoder that adds a 3D positional encoding to each image feature. This enhancement helps to effectively fuse information about objects that appear in multiple cameras. Focal-PETR \cite{focalpetrwang} proposes focal sampling and spatial alignment modules to adaptively focus on foreground regions. Another state-of-the-art sparse object-centric method is StreamPETR \cite{streampetrwang2023}, which extends Focal-PETR \cite{focalpetrwang} into a temporal object detector by introducing an efficient temporal object query propagation mechanism.
\subsection{Neural Rendering of Street View Scenes}
Neural rendering methods enable the generation of photorealistic images from novel viewpoints based on the 3D representation of the scene. Unlike prior methods with 3D supervision \cite{deepSDF, occupancyNetworks}, the Neural Radiance Field (NeRF) \cite{nerfmildenhall2021} continuous scene representation can be trained using posed 2D images alone. This flexibility is achieved by the differentiable volume rendering procedure, which turns view synthesis into a self-supervised learning problem. In the following, we focus on research aspects of NeRFs for reconstructing autonomous driving scenes, which is particularly challenging due to the dynamic nature and highly correlated views of driving sequences.
\subsubsection{Depth Supervision}
LiDAR point clouds, frequently available in autonomous driving datasets like \cite{nuscenescaesar2020, waymosun2020, kittiliao2022, oncemao2021, argoverse2wilson2023}, provide precise depth measurements and facilitate the integration of sparse depth supervision into NeRF optimization \cite{urbanRadianceFieldsrematas2022}. Previous research has shown that depth supervision can significantly improve the 3D reconstruction quality of street view scenes \cite{neuradtonderski2024, emernerfyang2023, unisimyang2023, streetsurfguo2023}, making it a common choice for neural rendering models in the autonomous driving domain.
\begin{table}[t!]
    \setlength{\tabcolsep}{5pt} 
    \centering
    \begin{tabular}{lcccc}
        \toprule
        \multicolumn{1}{l}{\multirow{2}{*}{\diagbox[width=8em]{\multirow{2}{*}{Method}}{Type}}} & Sparse & \multicolumn{2}{c}{Dense BEV-based} & \multirow{2}{*}{Temporal} \\
        \cmidrule(lr){3-4}
        \multicolumn{1}{l}{} & Object-centric & Forward & Backward & \\
        \toprule
        DETR3D \cite{detr3dwang2022} & $\times$ & & & \\
        PETR \cite{petrliu2022} & $\times$ & & & \\
        StreamPETR \cite{streampetrwang2023} & $\times$ & & & $\times$ \\
        \midrule
        BEVDet \cite{bevdethuang2021} & & $\times$ & &  \\
        BEVDet4D \cite{bevdet4dhuang2022} & & $\times$ & & $\times$ \\
        \midrule
        BEVFormer-S \cite{bevformerLi2022} & & & $\times$ & \\
        BEVFormer \cite{bevformerLi2022} & & & $\times$ & $\times$ \\ 
        \bottomrule
    \end{tabular}
    \caption{Categorization of 3D object detector architectures according to \cite{bevnextli2024} and \cite{surveyVisionCentricBEVPerceptionma2024}.}
    \label{tab:3d_object_detection_architectures}
\end{table}
\subsubsection{Dynamic Scene Representation}
One technique suitable for street view scenes is \textit{implicit decomposition}, where neural rendering methods learn to decompose a dynamic scene into a static 3D scaffold and a sparse 4D dynamic overlay without requiring manual annotations. SUDS \cite{sudsscalableurbandynamicscenesturki2023} and EmerNeRF \cite{emernerfyang2023} implement this paradigm by utilizing both a static and a dynamic encoder, which are implemented in 3D and 4D multi-resolutional hash encodings \cite{instantneuralgraphicsmuller2022}, respectively. In contrast, \textit{explicit decomposition} methods explicitly separate the scene into a static environment and dynamic actors, requiring supervision such as tracked bounding boxes for each instance. 
NeuRAD \cite{neuradtonderski2024} separates actor features $\mathbf{v}_a$ from the environment features $\mathbf{v}_e$ on the encoder level and employs a shared decoder MLP that regresses an implicit geometry $s$ and an appearance feature $\bm{f}$ given either geometry feature $\mathbf{v}_e$ or $\mathbf{v}_a$. These appearance features are rendered along each ray, and a CNN upsampling network finally decodes them into an RGB image patch \cite{neuradtonderski2024}.
\begin{table}[t]
    \centering
    \setlength{\tabcolsep}{5pt} 
    \begin{tabular}{lclll}
        \toprule
        Split & Town & Properties of Environment & \# Scenes & $\sum$ \\
        \midrule
        \multirow{9}{*}{Train} & 01 & Small town with river & 40 & \multirow{9}{*}{700} \\
        & 02 & Residential and commercial areas & 40 & \\
        & 04 & Small town in mountains & 40 & \\
        & 05 & Urban large junctions & 90 & \\
        & 06 & Highways with entrances and exits & 40 & \\
        & 07& Rural area with narrow roads & 40 & \\
        & 10 & Large junctions and skyscrapers & 100 & \\
        & 12 & Large town, rural and urban & 180 & \\
        & 15 & Large map with many parking lots & 130 & \\
        \midrule
        \multirow{2}{*}{Val} & 03 & Large junctions and highway & 90 & \multirow{2}{*}{150} \\
        & 13 & Large town, rural and urban & 60 & \\
        \bottomrule
    \end{tabular}
    \caption{Overview of the training and validation scenes in our CamShift dataset.}
    \label{tab:overview_of_training_and_validation_towns}
\end{table}
\subsection{Sensor Adaptation Benchmarking}
To evaluate the effectiveness of domain adaptation for perception models, many approaches \cite{trainInGermanyTestinUSAwang2020, st3dyang, st3d++yang, uni3dzhang, resimadzhang2024} rely on cross-dataset experiments, e.g. based on nuScenes \cite{nuscenescaesar2020}, Waymo Open Dataset \cite{waymosun2020}, KITTI-360 \cite{kittiliao2022} or ONCE \cite{oncemao2021}. Since there is no dataset publicly available that includes two or more redundant and complete camera sensor setups simultaneously capturing the surrounding environment from different vertical viewpoints, existing studies inevitably include both sensor-related and environment-related domain gaps. The most suitable real-world dataset for exclusively examining sensor adaptation is the recently proposed WayveScenes101 dataset \cite{wayvescenes101zurn2024}, which includes 101 diverse driving scenarios for novel view synthesis. However, WayveScenes101 includes four cameras designated for training and only a single forward-facing camera for recording off-axis validation images, making it unsuitable for examining sensor adaptation of a complete surrounding view camera sensor setup to a different vertical viewpoint. In contrast to real-world datasets, many synthetic autonomous driving datasets are specifically designed to examine domain adaptation effects \cite{iddadatasetalberti2020, shiftdatasetsun2022,selmadatasettestolina2023, synpassBehindEveryDomainzhang2024}. Sun \text{et al.} proposes SHIFT \cite{shiftdatasetsun2022}, a dataset with both discrete and continuous domain shifts but thereby not including sensor shifts. The IDDA \cite{iddadatasetalberti2020} dataset contains data recordings from five different viewpoints, but only captures the front camera view \cite{selmadatasettestolina2023}. On the other hand, the SELMA dataset \cite{selmadatasettestolina2023} includes recordings from surrounding cameras, yet only the front camera is equipped with a redundant sensor setup at a different height.
\section{SENSOR ADAPTATION BENCHMARK}
We propose a novel sensor adaptation benchmark to investigate how a cross-sensor domain gap affects the accuracy of 3D perception in autonomous driving. Since there is no real-world or synthetic dataset available that contains sensor data for at least two different vehicle types recorded simultaneously, we propose the CamShift dataset, simulated using CARLA \cite{carladosovitskiy2017}. In the following, we describe the dataset and the main objectives behind its generation.
\subsection{Cross-sensor Domain Gap Isolation}
To isolate and explore the cross-sensor domain gap, we generate all scenes using two different camera setups, simultaneously capturing the scene from different viewpoints but mounted on the same vehicle. This allows us to simulate the viewpoints of a sport utility vehicle (\texttt{SUV}) and a subcompact vehicle (\texttt{SUB}), resulting in the two splits \texttt{sim-SUV} and \texttt{sim-SUB}. Unlike previous cross-dataset studies \cite{semvecnetranganatha2024, trainInGermanyTestinUSAwang2020}, this ensures the same distribution of traffic scenarios, environmental conditions, and objects across both sensor setups, thereby enabling explicit and isolated investigation of the cross-sensor domain gap. The camera setups of CamShift closely mirrors the nuScenes \cite{nuscenescaesar2020} configuration, consisting of six cameras that provide a 360-degree field of view. We place the cameras of \texttt{sim-SUV} at realistic in-vehicle locations typical of an SUV. In the \texttt{sim-SUB} setup, cameras are mounted $\text{50}\,\si{cm}$ lower, with reduced distances between front and rear cameras by $\text{90}\,\si{cm}$, and between left and right cameras by $\text{20}\,\si{cm}$. The number of cameras and the camera intrinsics remain unchanged across both configurations. Figure \ref{fig:gt_sub} provides an example view from the front cameras that highlight the difference between \texttt{sim-SUV} and \texttt{sim-SUB}.
\subsection{Quality and Scale}
Large-scale datasets with sufficient diversity are essential for recent camera-based 3D object detectors to generalize well \cite{3dObjectDetectionForAutonomousDrivingAComprehensiveSurveymao20233}. Therefore, we align CamShift with the scale of nuScenes \cite{nuscenescaesar2020}, featuring 700 training and 150 validation scenes (see Table \ref{tab:overview_of_training_and_validation_towns}). 
\begin{table}[t]
    \centering
    \begin{tabular}{llllll}
        \toprule
        Category & \# Instances & Ratio [\%] & \multicolumn{3}{c}{\# Instances per Scene} \\
        \cmidrule(lr){4-6}
        & & & Total & Train & Val \\
        \midrule
        Ambulance & 226 & 0.9 & 0.3 & 0.3 & 0.3 \\
        Bicycle & 946 & 3.7 & 1.1 & 1.1 & 1.0 \\
        Bus & 294 & 1.1 & 0.3 & 0.3 & 0.4 \\
        Car & 15762 & 60.9 & 18.5 & 20.0 & 12.1 \\
        Human & 6114 & 23.6 & 7.2 & 7.0 & 8.1 \\
        Motorcycle & 1397 & 5.4 & 1.6 & 1.7 & 1.3 \\
        Truck & 1143 & 4.4 & 1.3 & 1.4 & 1.0 \\
        \bottomrule
    \end{tabular}
    \caption{Distribution of the seven object categories in our CamShift dataset.}
    \label{tab:object_category_distribution}
\end{table}
We use the CARLA traffic manager to create random yet realistic traffic scenes with high variability across all eleven towns, with the number of scenes per town based on the town's spatial extent and diversity (cf. Table \ref{tab:overview_of_training_and_validation_towns}). We adjust the weather conditions to cover different times of day, including night, dawn, and daytime, with varying levels of cloudiness and wind. Moreover, we use all available vehicle and pedestrian assets, vary the traffic density, and allow jaywalking for some pedestrians to increase the scene complexity. We employ a train-val partitioning strategy based on the location where the scene was recorded, in order to evaluate 3D object detection on independent static assets. Table \ref{tab:overview_of_training_and_validation_towns} shows the resulting split between the training and validation scenes, and is identical for both \texttt{sim-SUV} and \texttt{sim-SUB}. In Table \ref{tab:object_category_distribution}, the object category distribution is shown for all seven classes. Unlike nuScenes, our dataset does not contain annotations for the classes \textit{construction vehicle}, \textit{trailer}, \textit{barrier} and \textit{traffic cone}. This reduction is partially offset by the challenging class \textit{ambulance}, resulting in the two rare classes \textit{ambulance} and \textit{bus}, accounting for only 0.9\% and 1.1\% of all annotations, respectively.
\subsection{Drop-in Usability}
To streamline the cross-sensor domain gap evaluation of 3D object detectors, we store sensor data, annotations, and metadata generated by the CARLA simulator in a format that closely mimics the relational schema of the nuScenes dataset \cite{nuscenescaesar2020}, both syntactically and semantically. This allows CamShift to be used as a drop-in replacement for the official nuScenes dataset, minimizing the need to modify existing data loaders and evaluation scripts in the widely used \textit{mmdetection3d} \cite{mmdetection3d} framework.

In summary, we develop a challenging dataset that mimics common challenges in 3D object detection, such as class imbalance, occluded objects, and the detection of small objects. The direct correspondence between scenes in the \texttt{sim-SUV} and \texttt{sim-SUB} splits and the nuScenes drop-in usability facilitates sensor adaptation benchmarking.
\section{SENSOR ADAPTATION PIPELINE}
We propose a data-driven sensor adaptation pipeline based on neural rendering to convert entire scenes, and consequently whole autonomous driving datasets, from one sensor setup to another (see Figure \ref{fig:pipeline}). First, a 3D scene representation is learned based on the entire sequence of surrounding images. Then, to evaluate whether the quality of the learned representation is sufficient, images of the original viewpoint are rendered and compared to the original input images. If the minimum required quality is achieved, the images for the novel views of the new sensor setup are finally rendered. The pipeline can be used to change the number of cameras, their intrinsic parameters such as focal length or resolution, as well as their extrinsic parameters. However, we restrict our study to compensating for a pronounced sensor shift from an SUV platform to a subcompact platform matching our benchmark. To realize that, we have two constraints for the underlying neural rendering method. First, the choice of scene decomposition paradigm is crucial for accurately representing dynamic elements and rendering high-quality images. Second, previous work \cite{emernerfyang2023} has trained and evaluated neural rendering models on lower resolution images to manage the computational costs of high-resolution rendering of entire datasets. 

\begin{figure}[t]
    \centering
    \begingroup
    \footnotesize
    \includesvg[width=0.48\textwidth]{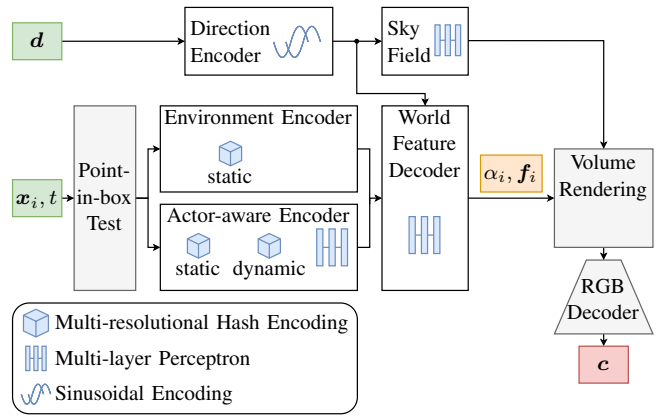}
    \endgroup
    \caption{Overview of our neural rendering architecture based on NeuRAD \cite{neuradtonderski2024}. For each sample point $\bm{x}_i$ at time $t$ either an environment or an actor-aware encoding is calculated. For actor-aware encodings, a small MLP determines the ratio between static and dynamic components. As the geometry and appearance decoders, adapted from NeuRAD \cite{neuradtonderski2024}, predict the world features, the sky field computes a sky feature solely based on the ray direction $\bm{d}$. Subsequent to volume rendering, a CNN-based RGB upsampling model decodes the color $\bm{c}$ from the sample features $\bm{f_i}$ and the opacities $\alpha_i$.}
    \label{fig:neural_rendering_model_architecture}
\end{figure}

Given these constraints, EmerNeRF \cite{emernerfyang2023}, which uses the implicit scene decomposition paradigm, and NeuRAD \cite{neuradtonderski2024}, which employs the explicit scene decomposition paradigm, can both serve as underlying method for our data-driven sensor adaptation pipeline. During initial experiments, we find that implicit scene decomposition tends to overfit to the training views, leading to degraded actor geometry when observed from a different camera setup. In contrast, explicit scene decomposition renders actors more accurately and consistently across different camera setups. Although explicit scene decomposition approaches like \cite{neuradtonderski2024, neuralscenegraphsost2021} require 3D bounding box annotations, we can reasonably assume that such annotations are available in autonomous driving datasets. Thus, we choose a neural rendering method based on NeuRAD \cite{neuradtonderski2024} as shown in Figure \ref{fig:neural_rendering_model_architecture}. Our approach is an extension of NeuRAD \cite{neuradtonderski2024} with the following improvements:
\begin{itemize}
    \item Explicit scene decomposition capable of modeling time-varying appearance of actors.
    \item Integration of a separate sky branch following \cite{emernerfyang2023, streetsurfguo2023} with enhanced supervision.
\end{itemize}
\begin{figure*}[t]
    \begingroup
    \small
    \centering
    \rotatebox[origin=left]{90}{\hspace{0.85cm}\texttt{SUV}}
    \begin{subfigure}[b]{0.20\textwidth}
        \centering
        \includegraphics[width=\textwidth]{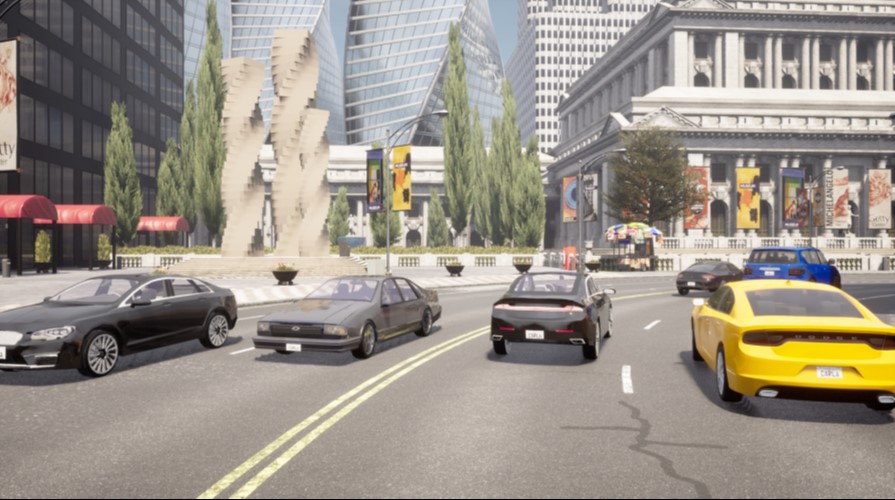}
        \label{fig:gt_suv}
    \end{subfigure}
    \begin{subfigure}[b]{0.20\textwidth}
        \centering
        \includegraphics[width=\textwidth]{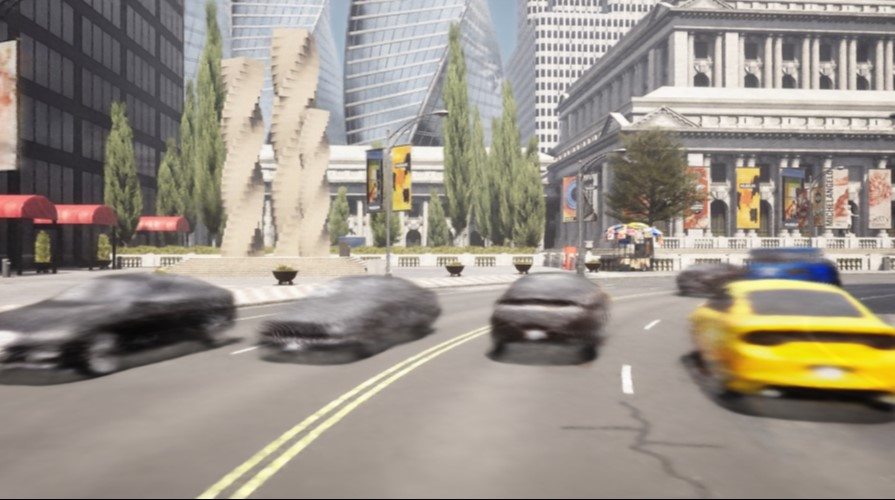}
        \label{fig:emernerf_suv}
    \end{subfigure}
    \begin{subfigure}[b]{0.20\textwidth}
        \centering
        \includegraphics[width=\textwidth]{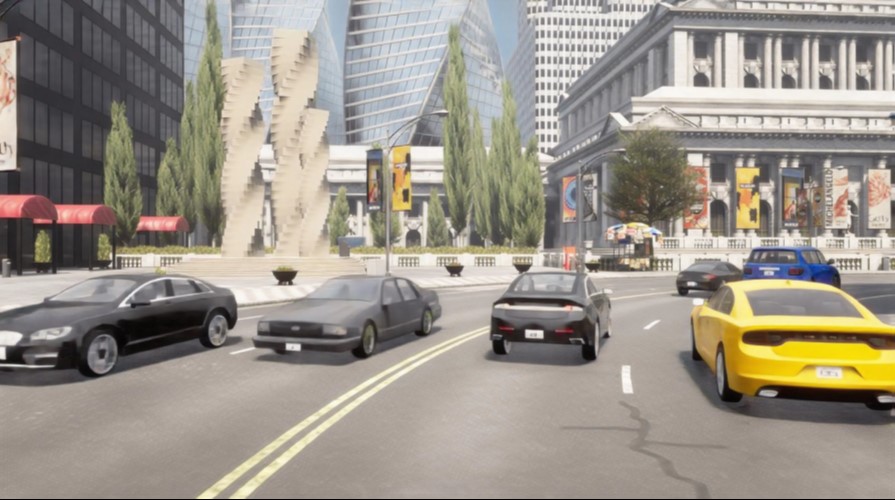}
        \label{fig:neurad_suv}
    \end{subfigure}
    \begin{subfigure}[b]{0.20\textwidth}
        \centering
        \includegraphics[width=\textwidth]{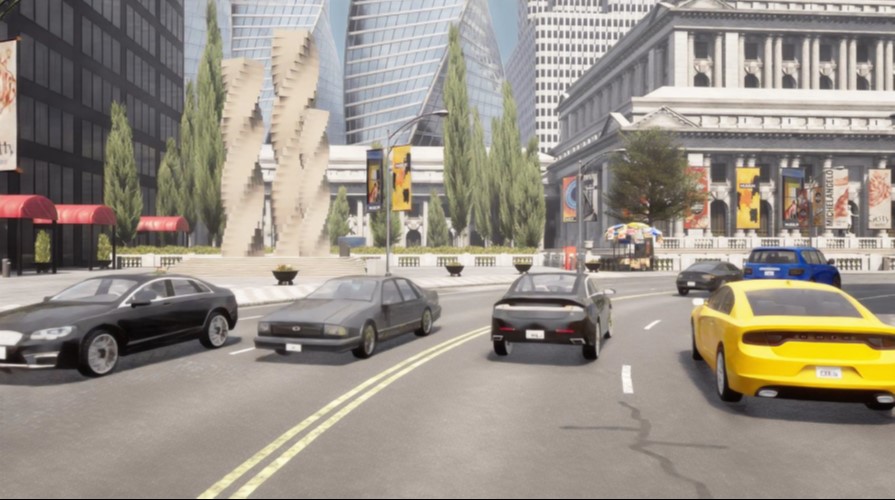}
        \label{fig:ours_suv}
    \end{subfigure}
    \\
    \vspace{-0.8\baselineskip}
    \rotatebox[origin=left]{90}{\hspace{1.1cm}\texttt{SUB}}
    \begin{subfigure}[b]{0.20\textwidth}
        \centering
        \includegraphics[width=\textwidth]{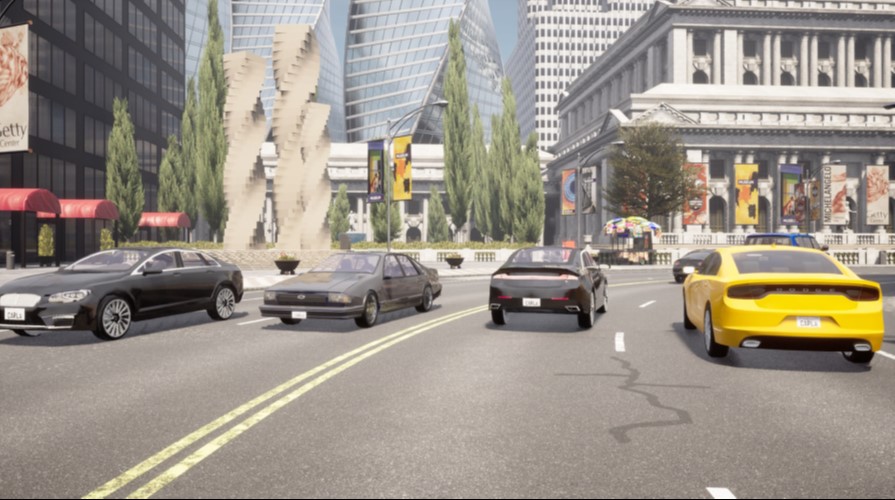}
        \caption{Ground Truth}
        \label{fig:gt_sub}
    \end{subfigure}
    \begin{subfigure}[b]{0.20\textwidth}
        \centering
        \includegraphics[width=\textwidth]{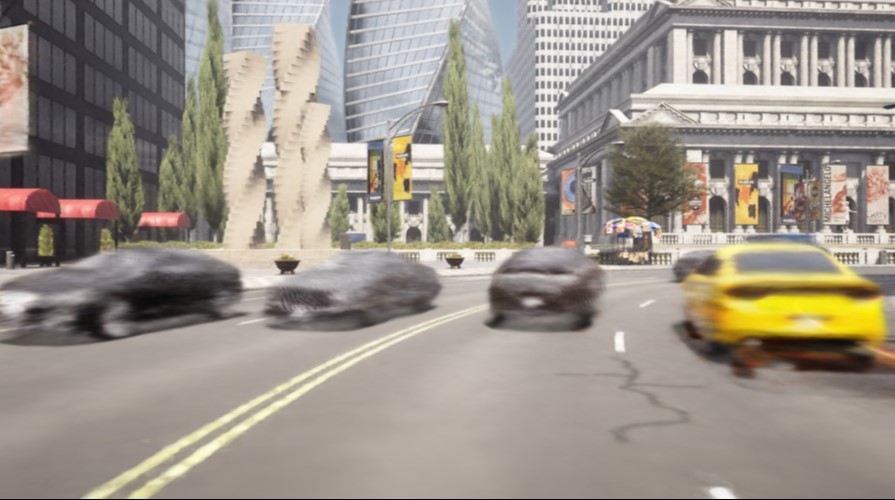}
        \caption{EmerNeRF \cite{emernerfyang2023}}
        \label{fig:emernerf_sub}
    \end{subfigure}
    \begin{subfigure}[b]{0.20\textwidth}
        \centering
        \includegraphics[width=\textwidth]{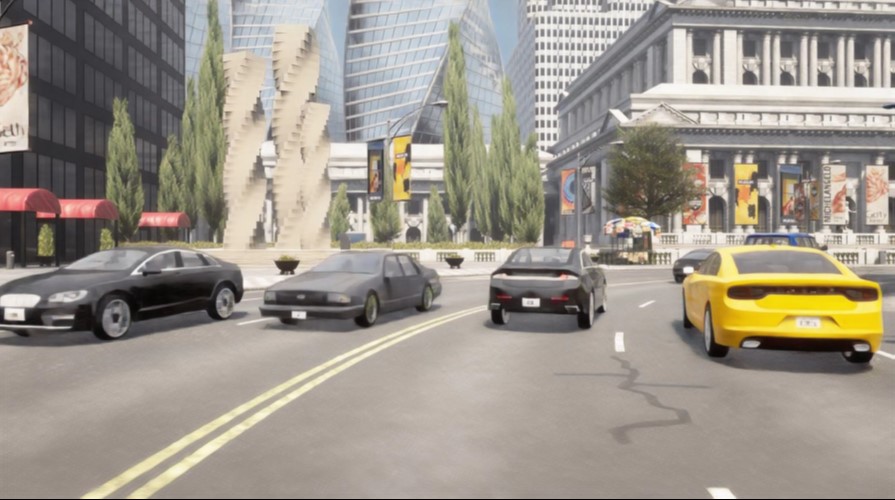}
        \caption{NeuRAD \cite{neuradtonderski2024}}
        \label{fig:neurad_sub}
    \end{subfigure}
    \begin{subfigure}[b]{0.20\textwidth}
        \centering
        \includegraphics[width=\textwidth]{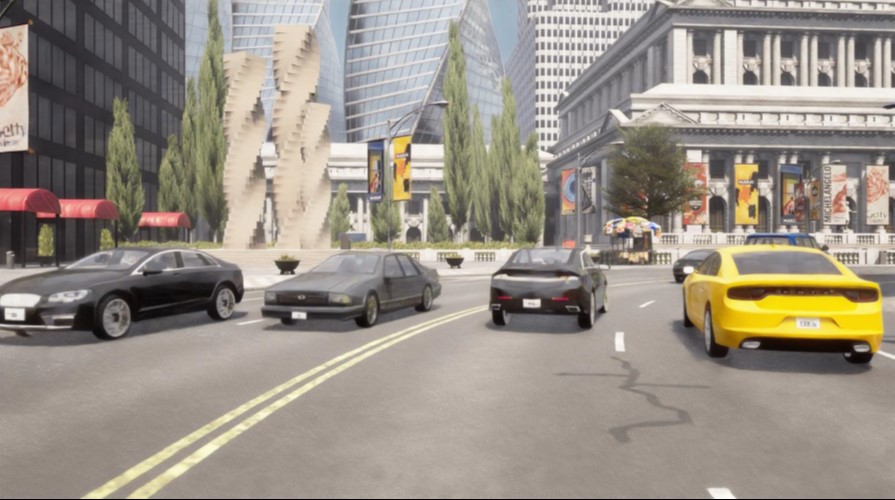}
        \caption{Ours}
        \label{fig:ours_sub}
    \end{subfigure}
    \caption{Qualitative comparison of the rendering quality for reconstructing the original view (\texttt{nerf-SUV}) and synthesizing the novel view (\texttt{nerf-SUB}) with two renowned neural rendering methods for autonomous driving scenes.}
    \label{fig:qualitative_comparison}
    \endgroup
\end{figure*}
\begin{table*}[h]
    \centering
    \begin{tabular}{lcc|ccc|ccc}
        \toprule
        & \multicolumn{2}{c|}{Speed} & \multicolumn{3}{c|}{Original View (\texttt{nerf-SUV})*} & \multicolumn{3}{c}{Novel View (\texttt{nerf-SUB})*} \\
        & Training $\downarrow$ & Rendering $\uparrow$ & PSNR $\uparrow$ & SSIM $\uparrow$ & LPIPS $\downarrow$ & PSNR $\uparrow$ & SSIM $\uparrow$ & LPIPS $\downarrow$ \\
        \midrule
        EmerNeRF \cite{emernerfyang2023} & $\text{2.61}\,\si{h}$ & 0.1 FPS & 27.68 & 0.741 & 0.364 & 24.58 & 0.663 & 0.424 \\
        NeuRAD \cite{neuradtonderski2024} & $\text{1.53}\,\si{h}$ & \textbf{7.6 FPS} & 27.98 & 0.744 & 0.170 & 25.12 & 0.668 & 0.235 \\
        Ours & $\textbf{1.49}\, \textbf{\si[detect-weight]{h}}$ & 3.6 FPS & \textbf{28.37} & \textbf{0.753} & \textbf{0.162} & \textbf{25.48} & \textbf{0.678} & \textbf{0.224} \\
        \bottomrule
    \end{tabular}
    \caption{Quantitative comparison of the rendering quality for reconstructing the original view (\texttt{nerf-SUV}) and synthesizing the novel view (\texttt{nerf-SUB}) with two renowned neural rendering methods for autonomous driving scenes.
    \\ * Metrics are calculated based on a small subset of \texttt{sim-SUV}.}
    \label{tab:image_quality_metrics}
\end{table*}
In Table \ref{tab:image_quality_metrics}, we compare our neural rendering model to both NeuRAD \cite{neuradtonderski2024} and EmerNeRF \cite{emernerfyang2023} based on training time per scene, rendering speed and the metrics for image quality assessment \cite{nerfmildenhall2021}, such as Peak Signal-to-Noise-Ratio (PSNR), Structural Similarity Index Measure (SSIM) \cite{imageQualityAssessmentwang2004} and Learned Perceptual Image Patch Similarity (LPIPS) \cite{UnreasonableEffectivenesszhang2018}. Due to limited computational resources, the results are based on a small subset of driving scenarios from \texttt{sim-SUV} at various times of day. We train the NeRFs on all camera images from the $\text{20}\,\si{Hz}$ scenes excluding the $\text{2}\,\si{Hz}$ key frames. We then render these excluded frames and validate them against the original images (see \emph{original view} column in Table \ref{tab:image_quality_metrics}). Since this common evaluation process \cite{unisimyang2023, neuradtonderski2024} involves only temporal interpolation, we extend the evaluation to include the rendering quality of novel viewpoints (see \emph{novel view} column in Table \ref{tab:image_quality_metrics}). This is crucial for the application in our sensor adaptation pipeline. Our method outperforms EmerNeRF and NeuRAD in terms of image quality in all metrics both for rendering orignal and novel view. NeuRAD achieves the second-best results, while EmerNeRF falls behind in terms of LPIPS. This is critical for 3D object detection, as LPIPS is a perceptual metric that closely correlates with human judgment of image quality \cite{UnreasonableEffectivenesszhang2018}, which can be also seen in Figure \ref{fig:qualitative_comparison}.
\section{SENSOR ADAPTATION EVALUATION}
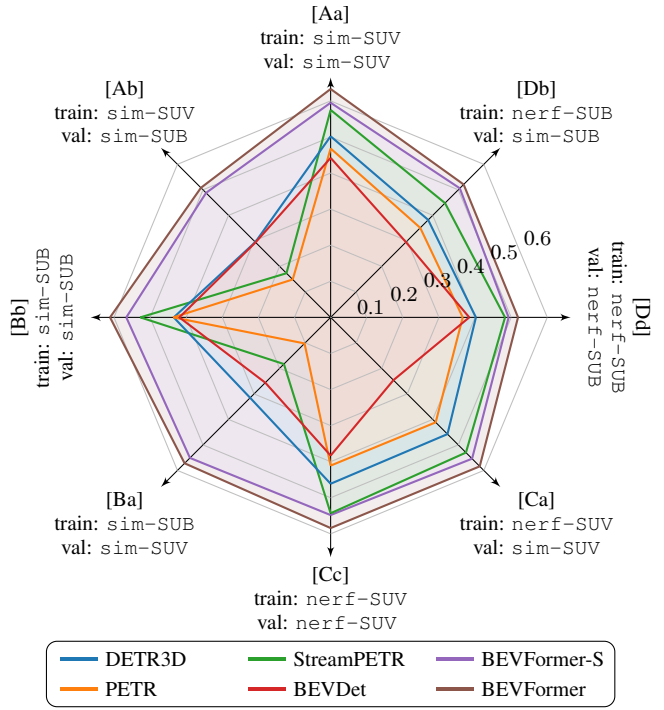
\begin{figure}[t!]
    \centering
    \begingroup
    \footnotesize
    \begin{tikzpicture}[trim left=-122pt]
\newcommand\ColorBox[1]{\textcolor{#1}{\rule{2ex}{2ex}}}

\tkzKiviatDiagram[scale=0.48,
        label space=7.85cm,
        radial  = 8,
        gap     = 1,  
        lattice = 6.15,
        rotate = 90]{[Aa] \\ train: \texttt{sim-SUV} \\ val: \texttt{sim-SUV} \vspace{-1em}, [Ab] \\ train: \texttt{sim-SUV} \\ val: \texttt{sim-SUB}, , [Ba] \\ train: \texttt{sim-SUB} \\ val: \texttt{sim-SUV}, [Cc] \\ train: \texttt{nerf-SUV} \\ val: \texttt{nerf-SUV} \vspace{1em}, [Ca] \\ train: \texttt{nerf-SUV} \\ val: \texttt{sim-SUV}, , [Db] \\ train: \texttt{nerf-SUB} \\ val: \texttt{sim-SUB}}

\node[rotate=90] at (0.0, 8.0) {\parbox{3cm}{\centering [Bb] \\ train: \texttt{sim-SUB} \\ val: \texttt{sim-SUB}}};
\node[rotate=-90] at (0.0, -8.0) {\parbox{3cm}{\centering [Dd] \\ train: \texttt{nerf-SUB} \\ val: \texttt{nerf-SUB}}};

\tkzKiviatLine[thick,color=mplbrown,mark=none,
               fill=mplbrown!20,opacity=.5](0.633*10,0.508*10,0.611*10,0.572*10,0.584*10,0.584*10,0.519*10,0.521*10)
\tkzKiviatLine[thick,color=mplpurple,mark=none,
               fill=mplpurple!20,opacity=.5](0.595*10,0.488*10,0.566*10,0.551*10,0.548*10,0.554*10,0.493*10,0.506*10)
\tkzKiviatLine[thick,color=mplgreen,mark=none,
               fill=mplgreen!20,opacity=.5](0.575*10,0.173*10,0.527*10,0.183*10,0.543*10,0.530*10,0.483*10,0.448*10)
\tkzKiviatLine[thick,color=mplblue,mark=none,
               fill=mplblue!20,opacity=.5](0.5019*10,0.2950*10,0.4346*10,0.3153*10,0.4612*10,0.4576*10,0.4024*10,0.3821*10)
\tkzKiviatLine[thick,color=mplorange,mark=none,
               fill=mplorange!20,opacity=.5](0.4681*10,0.1485*10,0.4357*10,0.1022*10,0.4109*10,0.4115*10,0.3659*10,0.3510*10)
\tkzKiviatLine[thick,color=mplred,mark=none,
               fill=mplred!20,opacity=.5](0.442*10,0.294*10,0.419*10,0.256*10,0.383*10,0.246*10,0.384*10,0.295*10)
               
\tkzKiviatGrad[prefix=0.,unity=1,suffix=, graduation distance=-15](6.5)

\node[xshift=0pt,yshift=-135pt, draw=black, fill=white, thin, inner sep=2pt, rounded corners] {
    \begin{tikzpicture}[baseline]
        \node[anchor=west] at (0-4.25, 0) {\textcolor{mplblue}{\rule{0.5cm}{0.05cm}}};
        \node[anchor=west] at (0.6-4.25, 0) {DETR3D};
        
        \node[anchor=west] at (0-4.25, -0.4) {\textcolor{mplorange}{\rule{0.5cm}{0.05cm}}};
        \node[anchor=west] at (0.6-4.25, -0.4) {PETR};
        
        \node[anchor=west] at (2.5-4.25, 0) {\textcolor{mplgreen}{\rule{0.5cm}{0.05cm}}};
        \node[anchor=west] at (3.1-4.25, 0) {StreamPETR};
        
        \node[anchor=west] at (2.5-4.25, -0.4) {\textcolor{mplred}{\rule{0.5cm}{0.05cm}}};
        \node[anchor=west] at (3.1-4.25, -0.4) {BEVDet};
        
        \node[anchor=west] at (5-4.25, -0) {\textcolor{mplpurple}{\rule{0.5cm}{0.05cm}}};
        \node[anchor=west] at (5.6-4.25, -0) {BEVFormer-S};
        
        \node[anchor=west] at (5-4.25, -0.4) {\textcolor{mplbrown}{\rule{0.5cm}{0.05cm}}};
        \node[anchor=west] at (5.6-4.25, -0.4) {BEVFormer};
    \end{tikzpicture}
};

\end{tikzpicture}
    \endgroup
    \vspace{-1em}
    \caption{Illustration of the mAP for various representative 3D object detectors. Vertical and horizontal axes represent baseline experiments on \texttt{sim-SUV} and \texttt{sim-SUB}, respectively. The diagonal axes on the left show the two experiments involving a cross-sensor domain gap, while those on the right illustrate the experiments with neural rendering domain gap.}
    \label{fig:radar_chart}
\end{figure}

We present a systematic evaluation scheme (see Table \ref{tab:evaluation_scheme}) for both the cross-sensor domain gap of 3D object detectors and the effectiveness of our proposed sensor adaptation pipeline. While \texttt{sim-SUV} and \texttt{sim-SUB} are the introduced platform splits of CamShift, \texttt{nerf-SUV} is the dataset of the rendered original views, generated using the proposed model shown in Figure \ref{fig:neural_rendering_model_architecture} and \texttt{nerf-SUB} are the scenes synthesized from \texttt{sim-SUV} for the novel views of a subcompact vehicle. On the one hand, with the experiments [Ba] and [Ab], we investigate the extent to which the sensor shift affects 3D object detectors compared to the respective baselines [Aa] and [Bb]. On the other hand, with experiment [Db], we examine the effectivity of our proposed data-driven sensor adaptation pipeline to mitigate the cross-sensor domain gap and reduce the need to collect and annotate additional training data for the new sensor setup. In Table \ref{tab:map} the mean average precision (mAP) for various representative 3D object detectors are summarized for all eight experiments shown in Table \ref{tab:evaluation_scheme}. Additionally, these mAP values are illustrated in the radar chart in Figure \ref{fig:radar_chart}. Table \ref{tab:mate} highlights the mean average translation error (mATE) for the three main experiments.

The implementations of all models are based on their latest version for the widely used \textit{mmdetection3d} framework \cite{mmdetection3d}. To ensure a fair comparison, all models are based on a ResNet-50 backbone \cite{resnethe2016} pretrained on ImageNet \cite{imagenetdeng2009} and the official model configurations and evaluation scheme from the nuScenes detection benchmark \cite{nuscenescaesar2020} were applied, with the only variation being in the classes (see Table \ref{tab:object_category_distribution}). By comparing the different experiments conducted for the selected 3D object detectors, we have identified the following key findings.
\paragraph{The 3D object detection performance depends on the choice of the camera setup}
The baseline experiments [Aa] and [Bb] consistently show across all detectors that detection performance depends on the camera setup. The mAP for the \texttt{SUV} platform is on average 3.4\% higher than for the \texttt{SUB} platform. This can be intuitively explained by Figure \ref{fig:gt_sub}, where the yellow car causes more significant occlusion in the subcompact sensor setup. Objects on the street are in general more likely to be occluded when recorded from a lower camera position, making it difficult to fully perceive the surroundings.

\paragraph{The cross-sensor domain gap significantly degrades the performance of camera-based 3D object detectors, and their robustness depends on the model architecture}
In the experiments [Ba] and [Ab], involving a cross-sensor
domain gap, the detection accuracy drops by 19.6\% mAP
on average across all detectors compared to the corresponding
baselines [Aa] and [Bb]. The performance degradation is approximately symmetric for the scenarios [Ba] and [Ab] which confirms, that neither setup is inherently better or worse for \emph{training} a detector. Instead, the performance drop results from overfitting to one specific camera setup. The degradation correlates with an increase in true positive error metrics, particularly the mATE increases by 57.5\% ($+\text{0.34}\,\si{m}$ mATE) on average when cross-validating on \texttt{sim-SUB} (see Table \ref{tab:mate}). The rise in mATE indicates that all detectors struggle, to varying extents, to accurately locate objects with a different camera setup.
\begin{table}[t!]
    \setlength{\tabcolsep}{1.25pt} 
    \centering
    \begin{tabular}{cccccc}
        \toprule
        \multirow{2}{*}{\textbf{Training}} & \multicolumn{5}{c}{\textbf{Validation Split}} \\
        \cmidrule(lr){2-6}
        \multicolumn{2}{c}{\multirow{2}{*}{\textbf{Split}}} & \texttt{sim-SUV} & \texttt{sim-SUB} & \texttt{nerf-SUV} & \texttt{nerf-SUB} \\
        \multicolumn{2}{c}{} & [a] & [b] & [c] & [d] \\
        \midrule
        \texttt{sim-SUV} & & \multirow{2}{*}{SUV baseline} & cross-sensor & \multirow{2}{*}{n/a} & \multirow{2}{*}{n/a} \\
        \text{[A]} & & & domain gap & & \\
        \cmidrule(lr){2-6}
        \texttt{sim-SUB} & & cross-sensor & \multirow{2}{*}{SUB baseline} & \multirow{2}{*}{n/a} & \multirow{2}{*}{n/a} \\
        \text{[B]} & & domain gap & & & \\
        \cmidrule(lr){2-6}
        \texttt{nerf-SUV} & & neural rendering & \multirow{2}{*}{n/a} & original view & \multirow{2}{*}{n/a} \\
        \text{[C]} &  & domain gap & & rendering & \\
        \cmidrule(lr){2-6}
        \texttt{nerf-SUB} & & \multirow{2}{*}{n/a} & neural rendering & \multirow{2}{*}{n/a} & novel view \\
        \text{[D]} & & & sensor adaptation & & rendering \\
        \bottomrule
    \end{tabular}
    \caption{3D object detection evaluation scheme: Overview of baselines and domain gaps. Cells marked with 'n/a' are excluded from evaluation as the experiments are either irrelevant or do not provide additional insights.}
    \label{tab:evaluation_scheme}
\end{table}

BEVFormer \cite{bevformerLi2022} demonstrates the highest robustness to the cross-sensor domain gap, with an average drop of only 7.2\% mAP for the static and temporal variants. This observation is consistent with the findings of the noisy camera extrinsic ablation studies reported in \cite{bevformerLi2022}. In Figure 4, the high robustness of both BEVFormer variants is evident from the well-filled left-side portions, which do not show any significant declines. The superior performance of BEVFormer is also demonstrated in experiment [Ab] in Table \ref{tab:mate}, as the increase in translation error is significantly less compared to the remaining detectors. We assume that the reason for the robustness lies in the model architecture. As a dense BEV-based method with backward projection, BEVFormer mainly operates in a fixed grid space. The update of the BEV queries is always applied to a fixed position using the explicit backward projection, making it unaware of the camera position. In addition, BEVFormer samples 2D image features around the projected reference points, further increasing its robustness. In contrast to BEVFormer, BEVDet \cite{bevdet4dhuang2022} lifts the 2D image features into the BEV space in a forward projection manner. This forward projection seems to be more sensitive to the cross-sensor domain gap, as the mAP drops are 12.5\% and 18.6\% for the \texttt{SUB} and the \texttt{SUV} platform, respectively.

\begin{table*}[t!]
    \setlength{\tabcolsep}{5pt} 
    \centering
    \begin{tabular}{lllllllll}
        \toprule
        \multirow{2}{*}{Exp.} & \multirow{2}{*}{Training} & \multirow{2}{*}{Validation} & \multicolumn{6}{c}{mAP $\uparrow [\%]$} \\
        \cmidrule(lr){4-9}
        & & & DETR3D \cite{detr3dwang2022}$^\dagger$ & PETR \cite{petrliu2022}$^\ddagger$ & StreamPETR \cite{streampetrwang2023}$^\ddagger$ & BEVDet \cite{bevdethuang2021}$^\ddagger$ & BEVFormer-S \cite{bevformerLi2022}$^\dagger$ & BEVFormer \cite{bevformerLi2022}$^\dagger$ \\
        \midrule
        Aa & \texttt{sim-SUV} & \texttt{sim-SUV} & 51.6 & 46.8 & 57.5 & 44.2 & 59.5 & 63.3 \\
        \textbf{Bb} & \textbf{\texttt{sim-SUB}} & \textbf{\texttt{sim-SUB}} & \textbf{46.1} & \textbf{43.6} & \textbf{52.7} & \textbf{41.9} & \textbf{56.6} & \textbf{61.1} \\
        Cc & \texttt{nerf-SUV} & \texttt{nerf-SUV} & 48.1 & 41.1 & 54.3 & 38.3 & 54.8 & 58.4 \\
        Dd & \texttt{nerf-SUB} & \texttt{nerf-SUB} & 44.0 & 36.6 & 48.3 & 38.4 & 49.3 & 51.9 \\
        \textbf{Ab} & \textbf{\texttt{sim-SUV}} & \textbf{\texttt{sim-SUB}} & \textbf{29.7 ($-$16.4)} & \textbf{14.9 ($-$28.7)} & \textbf{17.3 ($-$35.4)} & \textbf{29.4 ($-$12.5)} & \textbf{48.8 ($-$7.8)} & \textbf{50.8 ($-$10.3)} \\
        Ba & \texttt{sim-SUB} & \texttt{sim-SUV} & 32.0 & 10.2 & 18.3 & 25.6 & 55.1 & 57.2 \\
        Ca & \texttt{nerf-SUV} & \texttt{sim-SUV} & 47.7 & 41.2 & 53.0 & 24.6 & 55.4 & 58.4 \\
        \textbf{Db} & \textbf{\texttt{nerf-SUB}} & \textbf{\texttt{sim-SUB}} & \textbf{43.1 ($+$13.4)} & \textbf{35.1 ($+$20.2)} & \textbf{44.8 ($+$27.5)} & \textbf{29.5 ($+$0.1)} & \textbf{50.6 ($+$1.8)} & \textbf{52.1 ($+$1.3)} \\
        \bottomrule
    \end{tabular}
    \caption{The mean average precision (mAP) for various representative 3D object detectors throughout all experiments. The values in parentheses highlight the cross-sensor domain gap and the corresponding performance gain achieved by our data-driven sensor adaptation pipeline for the \texttt{SUB} platform. The three main experiments are highlighted in bold. \\ $^\dagger$ $[\text{1600} \times \text{900}]$ input resolution. $^\ddagger$ $[\text{1408} \times \text{512}]$ input resolution.
    }
    \label{tab:map}
\end{table*}
\begin{table*}[t!]
    \setlength{\tabcolsep}{5pt} 
    \centering
    \begin{tabular}{lllllllll}
        \toprule
        \multirow{2}{*}{Exp.} & \multirow{2}{*}{Training} & \multirow{2}{*}{Validation} & \multicolumn{6}{c}{mATE $\downarrow$ [\si{m}]} \\
        \cmidrule(lr){4-9}
        & & & DETR3D \cite{detr3dwang2022}$^\dagger$ & PETR \cite{petrliu2022}$^\ddagger$ & StreamPETR \cite{streampetrwang2023}$^\ddagger$ & BEVDet \cite{bevdethuang2021}$^\ddagger$ & BEVFormer-S \cite{bevformerLi2022}$^\dagger$ & BEVFormer \cite{bevformerLi2022}$^\dagger$ \\
        \midrule
        {Bb} & \texttt{sim-SUB} & \texttt{sim-SUB} & {0.67} & {0.74} & {0.61} & {0.62} & {0.49} & {0.45} \\
        {Ab} & \texttt{sim-SUV} & \texttt{sim-SUB} & {0.89 ($+$0.22)} & {1.25 ($+$0.51)} & {1.22 ($+$0.61)} & {1.05 ($+$0.43)} & {0.62 ($+$0.13)} & {0.61 ($+$0.16)} \\
        {Db} & \texttt{nerf-SUB} & \texttt{sim-SUB} & {0.74 ($-$0.15)} & {0.88 ($-$0.37)} & {0.63 ($-$0.59)} & {0.77 ($-$0.28)} & {0.54 ($-$0.08)} & {0.53 ($-$0.08)} \\
        \bottomrule
    \end{tabular}
    \caption{The mean average translation error (mATE) for various representative 3D object detectors in the three main experiments. The values in parentheses highlight the cross-sensor domain gap and the corresponding performance gain achieved by our data-driven sensor adaptation pipeline for the \texttt{SUB} platform. \\ $^\dagger$ $[\text{1600} \times \text{900}]$ input resolution. $^\ddagger$ $[\text{1408} \times \text{512}]$ input resolution.
    }
    \label{tab:mate}
\end{table*}
The cross-sensor domain gap significantly degrades the detection accuracy of sparse object-centric methods, even if they primarily operate in 3D space. This is evident in Figure 4, where the left-side regions of DETR3D, PETR, and StreamPETR show substantial declines in the diagonal cross-sensor values. The reason for the large performance gaps lies in the update process of the sparse object-centric queries. In contrast to BEV queries lying in a fixed grid, sparse object queries implicitly encode a 3D position, which can change through the update process of the queries in the transformer layers. Consequently, these update steps tend to overfit on the sensor setup of the training dataset, reducing the performance on the new sensor configuration of the validation dataset. DETR3D \cite{detr3dwang2022} shows an average mAP reduction of 18.0\%, whereas PETR \cite{petrliu2022} experiences a more pronounced drop of 32.7\%. The 3D position-aware features, introduced by PETR, are intended to improve object localization but inadvertently cause even more overfitting to a specific camera setup. This is also demonstrated by a large cross-sensor translation error increase of 68.9\% ($+\text{0.51}\,\si{m}$ mATE), as shown in Table \ref{tab:mate}.

The temporal methods StreamPETR \cite{streampetrwang2023} and BEVFormer \cite{bevformerLi2022} are slightly more affected by the domain gap than their non-temporal counterparts, indicating that the accumulation of temporal features does not positively impact the cross-sensor robustness. Instead, StreamPETR experiences the highest degradation with an mAP drop of 37.3\% on average.

In summary, the robustness of 3D object detectors benefits from primarily operating in a camera-invariant 3D space, utilizing fixed grid-shaped BEV queries and backward projection to aggregate 2D image features. In contrast, sparse object-centric methods tend to overfit to the camera setup of the training data, and the aggregation of temporal information does not mitigate cross-sensor degradation.

\paragraph{Neural rendered images are suitable for 3D object detection}
By comparing the neural rendering baselines [Cc] and [Dd] with the corresponding simulation baselines [Aa] and [Bb], we find that neural rendering decreases the mAP on average by 5.1\% across all detectors. This is expected, as neural rendering cannot perfectly reconstruct the original images but introduces artifacts and reduces image details. Notably, the performance degradation is slightly more pronounced for the \texttt{SUB} platform ($-\text{5.6}$\% mAP) than for the \texttt{SUV} platform ($-\text{4.7}$\% mAP). This is consistent with the image quality metrics from Table \ref{tab:image_quality_metrics}, where the metrics are lower for the novel views of \texttt{sim-SUB}. Nevertheless, the reduction in detection metrics remains within acceptable limits and is consistently lower than cross-sensor degradation, which indicates that neural rendered images are suitable for perception tasks in general. Experiment [Ca] shows that a detector can be trained with neural rendered images and evaluated with original sensor data. Although this setting is unrealistic, it motivates the use for our sensor adaptation pipeline. The overall suitability of neural rendered images is also demonstrated in Figure \ref{fig:radar_chart} by comparing the right with the corresponding left segments.
\paragraph{Neural rendering mitigates the cross-sensor domain gap in 3D object detection}
Compared to the uncompensated experiment [Ab], applying our sensor adaptation pipeline [Db] has a huge impact on the 3D object detection performance, increasing the mAP by an average of 10.7\% across all detectors. In other words, we can mitigate the degradation of mAP from 18.5\% to just 7.8\% for the \texttt{SUB} platform, closing the gap by more than half. We observe the highest improvements for detectors that degrade the most due to the sensor shift. The positive effect can be further explained by mATE, where our pipeline is able to reduce the degradation in localization on average from $\text{0.34}\,\si{m}$ to just $\text{0.09}\,\si{m}$ due to adjusting the perspectives. For StreamPETR, our pipeline is even able to reduce the gap to just $\text{0.02}\,\si{m}$ mATE, resulting in a significant performance gain ($+\text{27.5}$\% mAP). Moreover, it is noteworthy that our pipeline has a positive impact on all 3D object detectors, including the BEVFormer variants, where the cross-sensor degradation is smallest. This is also illustrated in Figure \ref{fig:radar_chart}, where BEVFormer and BEVFormer-S are the only methods that fill the entire circle without having any significant declines.
\section{CONCLUSION AND FUTURE WORK}
In this paper, we introduced a novel sensor adaptation benchmark that explicitly targets the cross-sensor domain gap, a challenge that inevitably arises when scaling 3D object perception to a fleet of heterogeneous vehicle types. Through this benchmark, we demonstrated that the cross-sensor domain gap significantly impacts 3D object detection performance, leading to notable degradations.

To address this issue, we identified and proposed two complementary mitigation strategies. First, we observed that dense BEV-based architectures with backward projection, such as BEVFormer \cite{bevformerLi2022}  exhibit significantly higher robustness to cross-sensor variations compared to sparse, query-based approaches. Second, independent of the model architecture, our proposed sensor adaptation pipeline enabled dataset transformation to new sensor setups by leveraging neural rendering to generate respective novel views.

By employing this data-driven approach, we successfully mitigated the cross-sensor domain gap, substantially improving model generalization across different sensor configurations. Our proposed sensor adaptation pipeline thus provides a scalable solution for enhancing data reusability, reducing the need for extensive new data collection, and facilitating the deployment of 3D object perception systems in diverse autonomous vehicle fleets.

Future work could adapt our pipeline to real-world datasets with additional challenges such as adverse weather conditions or the image synthesis for fish-eye and telecameras. Moreover, training 3D object detectors on heterogeneous datasets generated by our pipeline could potentially improve generalization across different vehicle types. Additionally, integrating methods to transfer LiDAR point clouds across sensor setups opens up an interesting research direction. Lastly, we anticipate that further progress in neural rendering, such as 3D gaussian splatting \cite{gaussianSplatting}, will amplify the impact of our sensor adaptation pipeline.

\section*{ACKNOWLEDGEMENT}
This work is a result of the joint research project STADT:up (19A22006O). The project is supported by the German Federal Ministry for Economic Affairs and Climate Action (BMWK), based on a decision of the German Bundestag. The author is solely responsible for the content of this publication.

\bibliographystyle{ieeetr}
\bibliography{references/bibliography_short, references/short_references_sensor_adaptation}

\end{document}